\documentclass[conference]{IEEEtran}
\usepackage{times}

\usepackage[numbers]{natbib}
\usepackage{multicol}
\usepackage[bookmarks=true]{hyperref}

\usepackage{graphicx}%
\usepackage{multirow}%
\usepackage{amsmath,amssymb,amsfonts}%
\usepackage{amsthm}%
\usepackage{mathrsfs}%
\usepackage{xcolor}%
\usepackage{textcomp}%
\usepackage{manyfoot}%
\usepackage{booktabs}%
\usepackage{algorithm}%
\usepackage{algorithmicx}%
\usepackage{algpseudocode}%
\usepackage{listings}%
\usepackage{comment}
\usepackage[export]{adjustbox}
\usepackage{placeins}
\usepackage{caption}
\usepackage{cuted}

\begin{document}

\title{Supervised Mixture-of-Experts for Surgical Grasping and Retraction}

\author{\authorblockN{Lorenzo Mazza\textsuperscript{$\dagger$*1,2,3},
Ariel Rodriguez\textsuperscript{$\dagger$*1,2,3},
Rayan Younis\textsuperscript{4,5},
Martin Lelis\textsuperscript{1,2,3},\\
Ortrun Hellig\textsuperscript{4,5},
Chenpan Li\textsuperscript{1,2},
Sebastian Bodenstedt\textsuperscript{1,2,3,4},
Martin Wagner\textsuperscript{4,5},
Stefanie Speidel\textsuperscript{1,2,3,4}} \\
\authorblockA{\textsuperscript{$\dagger$}Equal contribution}
\authorblockA{\textsuperscript{1}Department of Translational Surgical Oncology, NCT/UCC Dresden, \\Faculty of Medicine and University Hospital Carl Gustav Carus, TUD, Dresden, Germany}
\authorblockA{\textsuperscript{2}National Center for Tumor Diseases (NCT), NCT/UCC Dresden, a partnership between DKFZ, \\Faculty of Medicine and University Hospital Carl Gustav Carus, TUD, and HZDR, Dresden, Germany.}
\authorblockA{\textsuperscript{3}Faculty of Computer Science, TUD, Dresden, Germany.}
\authorblockA{\textsuperscript{4}The Center for Tactile Internet with Human-in-the-Loop (CeTI), TUD, Dresden, Germany}
\authorblockA{\textsuperscript{5}Department of Visceral, Thoracic and Vascular Surgery, \\Faculty of Medicine and University Hospital Carl Gustav Carus, TUD, Dresden, Germany}
\authorblockA{\textsuperscript{*}Corresponding authors: lorenzo.mazza@nct-dresden.de, ariel.rodriguezjimenez@nct-dresden.de}
}



%

\maketitle
    
\begin{strip}
    \centering
    \includegraphics[width=\textwidth]{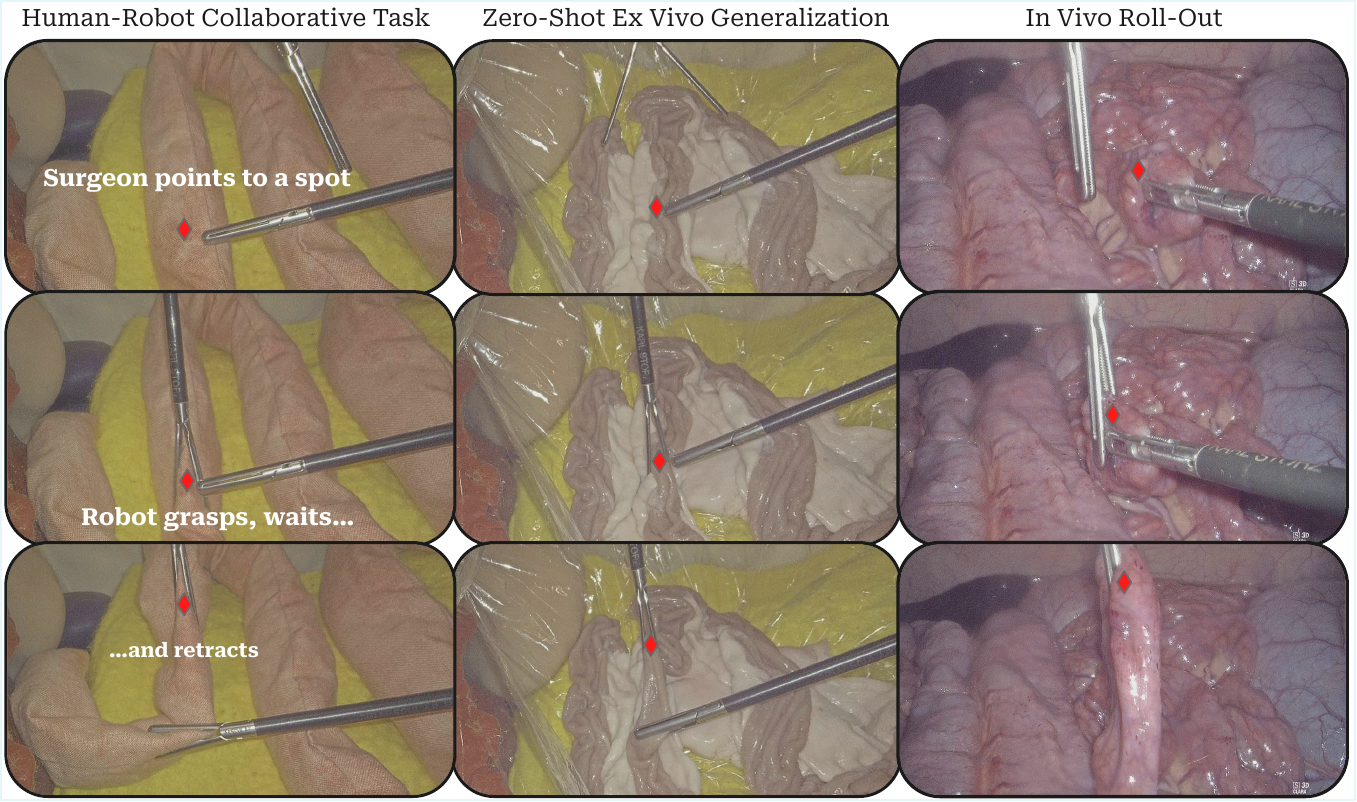}
    \captionof{figure}{Collaborative autonomous bowel grasping and retraction policy roll-outs across phantom, \textit{ex vivo}, and preliminary \textit{in vivo} porcine settings. The surgeon-controlled gripper (bottom instrument) indicates the target bowel segment, marked by the red rhomboid, while the robot assistant gripper (top instrument) autonomously performs grasping, retraction, and tension maintenance.}
    \label{fig_frontcover}
    \vspace{1em}
\end{strip}

\begin{abstract}
Imitation learning has achieved remarkable success in robotic manipulation, yet its application to surgical robotics remains challenging due to data scarcity, constrained workspaces, and the need for an exceptional level of safety and predictability. We present a supervised Mixture-of-Experts (MoE) architecture designed for phase-structured surgical manipulation tasks, which can be added on top of any autonomous policy. Unlike prior surgical robot learning approaches that rely on multi-camera setups or thousands of demonstrations, we show that a lightweight action decoder policy like Action Chunking Transformer (ACT) can learn complex, long-horizon manipulation from less than 150 demonstrations using solely stereo endoscopic images, when equipped with our architecture. We evaluate our approach on the collaborative surgical task of bowel grasping and retraction, where a robot assistant interprets visual cues from a human surgeon, executes targeted grasping on deformable tissue, and performs sustained retraction. 
We benchmark our method against state-of-the-art Vision-Language-Action (VLA) models and the standard ACT baseline. Our results show that generalist VLAs fail to acquire the task entirely, even under standard in-distribution conditions. Furthermore, while standard ACT achieves moderate success in-distribution, adopting a supervised MoE architecture significantly boosts its performance, yielding higher success rates in-distribution and demonstrating superior robustness in out-of-distribution scenarios, including novel grasp locations, reduced illumination, and partial occlusions.
Notably, it generalizes to unseen testing viewpoints and also transfers zero-shot to \textit{ex vivo} porcine tissue without additional training, offering a promising pathway toward \textit{in vivo} deployment. To support this statement, we present qualitative preliminary results of policy roll-outs during \textit{in vivo} porcine surgery. These results demonstrate that supervised MoE architectures provide a data-efficient approach for learning multi-step dexterous manipulation in visually constrained environments. Code and dataset will be released upon acceptance. \footnote{Available at \url{https://surgical-moe-project.github.io/rss-paper/}.}
\end{abstract}

\IEEEpeerreviewmaketitle
\section{Introduction}
Imitation learning (IL) has shown remarkable results in learning manipulation tasks through generative modeling approaches \cite{act, chi2023diffusion}. Recently, Vision-Language-Action (VLA) models \cite{black2024pi0, octo2024, kim2024openvla, pmlr-v305-black25a, pertsch2025fast, shukor2025smolvla, kim2025fine} have achieved impressive performance by leveraging large-scale datasets \cite{brohan2023rt1, openx2024, khazatsky2024droid}, demonstrating that foundation models can enable generalist robot policies across diverse tasks and embodiments. 

Minimally-invasive surgery (MIS) stands out as a particularly impactful application domain for autonomous manipulation. Staff shortages are expected to worsen relative to the growing surgical treatment needs of our ageing society worldwide \cite{perera2021global}, creating an urgent need for autonomous surgical assistance. Robot policies show great potential to address this shortcoming by enabling intraoperative autonomous assistance \cite{long2025surgical, younis2024surgical}, yet several challenges limit the direct adoption of general-purpose IL approaches.
Demonstration data is scarce due to ethical and regulatory constraints, inability to repeat procedures purely for data collection, and the prohibitive costs of operating room time and expert surgeon involvement. Data quality is further compromised by noise, due to occlusions, limited control over recording conditions, and the inherent variability of scenes in surgical procedures. Furthermore, workspace constraints preclude multi-view camera or depth sensor setups, tissue deformation adds complex dynamics not present in rigid object manipulation, and the proximity to delicate anatomical structures demands an exceptional level of safety and predictability. Surgical policies must additionally satisfy strict deployment requirements: lightweight architectures that fit on compact hardware with low inference latency to enable real-time control on resource-constrained systems. 

These constraints preclude the use of large pretrained VLA models, which i) require to be trained on large-scale datasets, due to their high parameter count and ii) incur high computational overhead during inference. Recent benchmarks in precision surgical tasks, such as end-to-end suturing \cite{haworth2025suturebot}, confirm that compact, lightweight policies like Action Chunking Transformers (ACT) \cite{act} outperform significantly VLAs when trained on limited surgical data, achieving considerably higher success rates and faster inference.

Similarly, Surgical Robotics Transformer (SRT) \cite{kim2024surgical} and its hierarchical variant SRT-H \cite{kim2025srt} have demonstrated that lightweight action transformer policies can learn dexterous multi-step surgical manipulation tasks from visual observations, proving that long-horizon precision tasks can be learned in a data-driven manner. However, key challenges remain unsolved. First, these approaches rely on multi-camera setups — including wrist-mounted cameras — to ensure robust 3D scene understanding, configurations that are often infeasible in MIS settings where only a single endoscopic view of the scene is available. Second, they still require extensive demonstration datasets: for instance, SRT-H's gallbladder clipping and cutting required approximately 16,000 demonstrations to achieve reliable performance.
To address these shortcomings, we propose a supervised Mixture-of-Experts (MoE) extension to action transformer policies. MoE architectures offer a promising framework for modeling multi-step, long-horizon surgical tasks by employing specialized sub-networks (experts) that handle different aspects of the task space \cite{jacobs1991adaptive, jordan1994hierarchical}. In robotics, MoE has shown promise for learning diverse skills and handling multi-modal action distributions \cite{reuss2024efficient, song2024germ}. 
The key insight is that complex tasks can be decomposed into simpler sub-components, each handled by a dedicated expert, with a gating mechanism determining which experts to activate based on the current context. This decomposition is particularly relevant for surgical tasks, where phase transitions are often well-defined and observable \cite{maier2017surgical, funke2019using}.
However, training MoEs end-to-end is notoriously unstable, often suffering from mode collapse or expert underutilization where the gating mechanism fails to effectively distribute task dynamics \cite{zhou2024variational}. We overcome these optimization challenges by exploiting the ordered phase structure of surgical sub-tasks, supervising explicitly the gating network with phase labels, ensuring stable convergence and clear functional specialization for each expert.
\\

Assistant tissue manipulation tasks in MIS mostly consist of tissue grasping and retraction, generally performed under the guidance of the operating surgeon \cite{collaborative2018perioperative, chiu2008role}. In MIS for gastrointestinal cancer treatment, surgical assistants must manipulate the small bowel to enable the surgeon to perform anastomoses on the jejunum and ileum \cite{konstantinidis2020trends}. This highlights the need to automate the assistant's role in small bowel manipulation, in particular bowel grasping and retraction, while maintaining coordinated interaction with the operating surgeon. 

In summary, our major contributions are the following:
\begin{itemize}
    \item  We propose a novel supervised Mixture of Experts (MoE) architecture designed for phase-structured surgical tasks that can be integrated into any kind of action transformer policies. We apply our architecture to a lightweight policy such as ACT and show that it can learn from significantly fewer demonstrations than prior work, and relying solely on endoscopic visual feedback — without wrist cameras or multi-view setups — for practical deployment in clinical MIS environments.
    \item We introduce a novel surgeon-robot collaboration task in laparoscopic bowel retraction, where a human surgeon provides high-level visual cues via a laparoscopic instrument, and the robot executes precise grasping, pulling, and sustained retraction actions. This cooperative paradigm emphasizes human-robot teamwork to enhance efficiency in MIS, where the robot serves as an intelligent assistant handling secondary but crucial tasks, i.e. maintains tissue retraction and tension, while the surgeon focus on critical actions — such as performing suturing or bowel anastomosis.
    \item We empirically validate the limitations of current state-of-the-art VLAs in the surgical domain. Our results reinforce recent findings that generalist foundation models fail to acquire high-precision surgical policies in data-scarce regimes, establishing the necessity of specialized, lightweight architectures for robust surgical automation. 
    \item We demonstrate two key prerequisite for \textit{in vivo} translation of autonomous surgical policies: (i) viewpoint invariance, showing that training with randomized camera angles enables the policy to generalize to unseen viewpoints without explicit 3D representations; and (ii) zero-shot transfer, where the policy achieves an 80\% success rate on \textit{ex vivo} porcine tissue despite being trained solely on phantom data. These results validate the system's robustness to the geometric variations and visual domain shifts inherent in dynamic clinical environments.    \end{itemize}
    Additionally, we release our code and dataset to facilitate reproducibility and future research in surgical robotics \footnote{Available at \url{https://surgical-moe-project.github.io/rss-paper/}.}.

\section{Methods}\label{sec_methods}
\subsection{Hardware and Experimental Setup}\label{subsec_hardware}

\begin{figure}[h]
    \centering
    \includegraphics[width=0.45\columnwidth, trim=2cm 10cm 0cm 2cm, clip, valign=c]{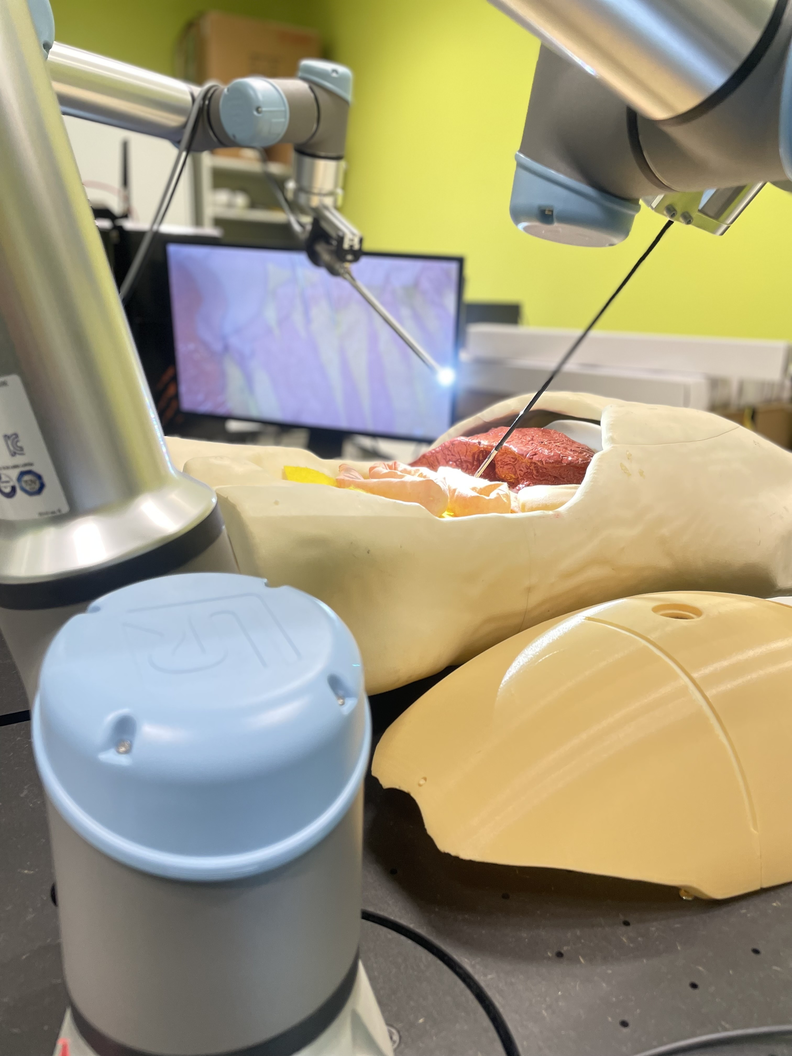}%
    \hspace{0.02\columnwidth}%
    \includegraphics[width=0.45\columnwidth, angle=-90, origin=c, trim=10cm 16cm 5cm 13cm, clip, valign=c]{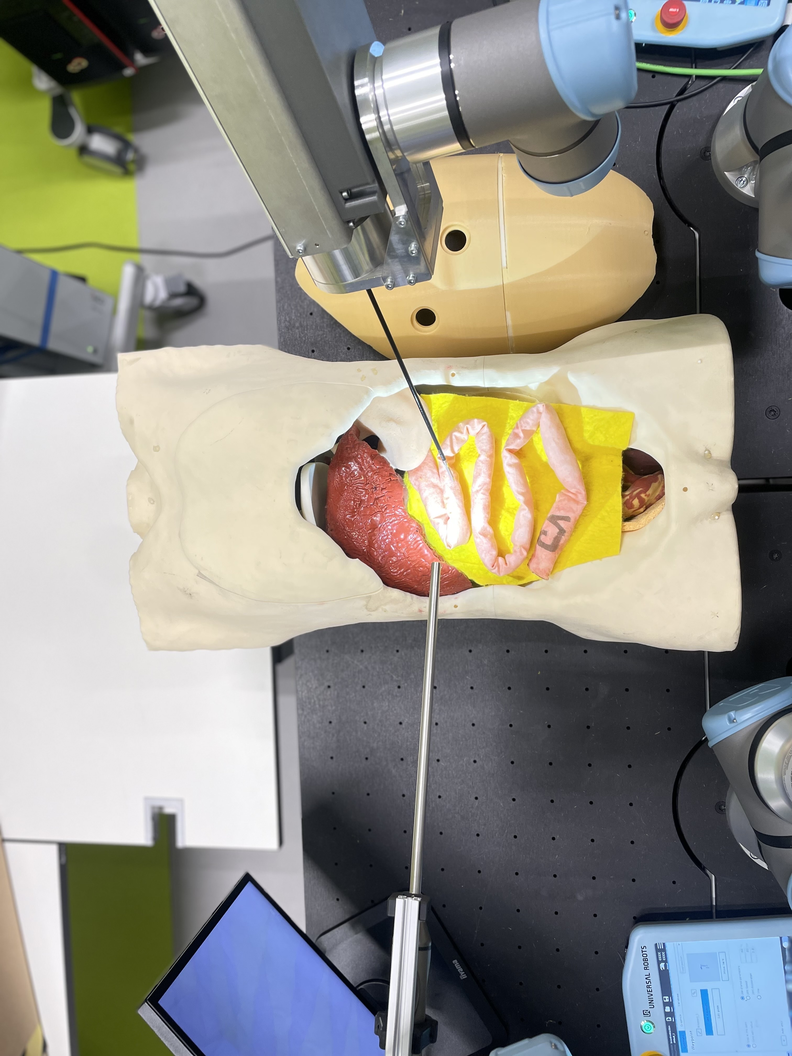}
    \caption{Experimental setup using the OpenHELP open-body phantom, showing the phantom with two robotic arms, one holding the laparoscope and one holding the surgical instrument. The abdominal wall cover is removed for visibility purposes.}
    \label{fig_environment}
\end{figure}

We develop an experimental setup using the OpenHELP open-body phantom \cite{openhelp}, as shown in figure \ref{fig_environment}.
We use two UR5e industrial robotic arms: one arm remains static and is equipped with a stereo TIPCAM1 S 3D endoscope (Karl Storz SE \& Co. KG) to provide visual feedback, while the other is equipped with a mechatronic interface \cite{schussler2025semi} that allows the attachment of a laparoscopic surgical bowel grasper, allowing controlled opening and closing of the gripper. The latter robot moves while maintaining the remote-center-of-motion (RCM)  constraint and is operated via a joystick-based control interface.

\subsection{Data Collection}\label{subsec_datacollection}
\begin{table}[h]
\centering
\caption{Task phase segmentation for bowel grasping and retraction.}
\label{tab_phases}
\begin{tabular}{@{}clp{4.5cm}@{}}
\toprule
Phase & Name & Description \\
\midrule
1 & Idle & Robotic instrument visible; surgeon's instrument may or may not be present. \\
2 & Approach \& Grasp & Surgeon indicates target; robot opens grasper, moves to target, and grasps tissue. \\
3 & Hold & Robot holds bowel stationary until surgeon grasps opposite end. \\
4 & Retract & Robot retracts bowel until visually straight and free of folds. \\
5 & Maintain Tension & Robot maintains tension with no further movement while surgeon operates. \\
\bottomrule
\end{tabular}
\end{table}
\begin{figure*}[t]
    \centering
    \includegraphics[width=\textwidth]{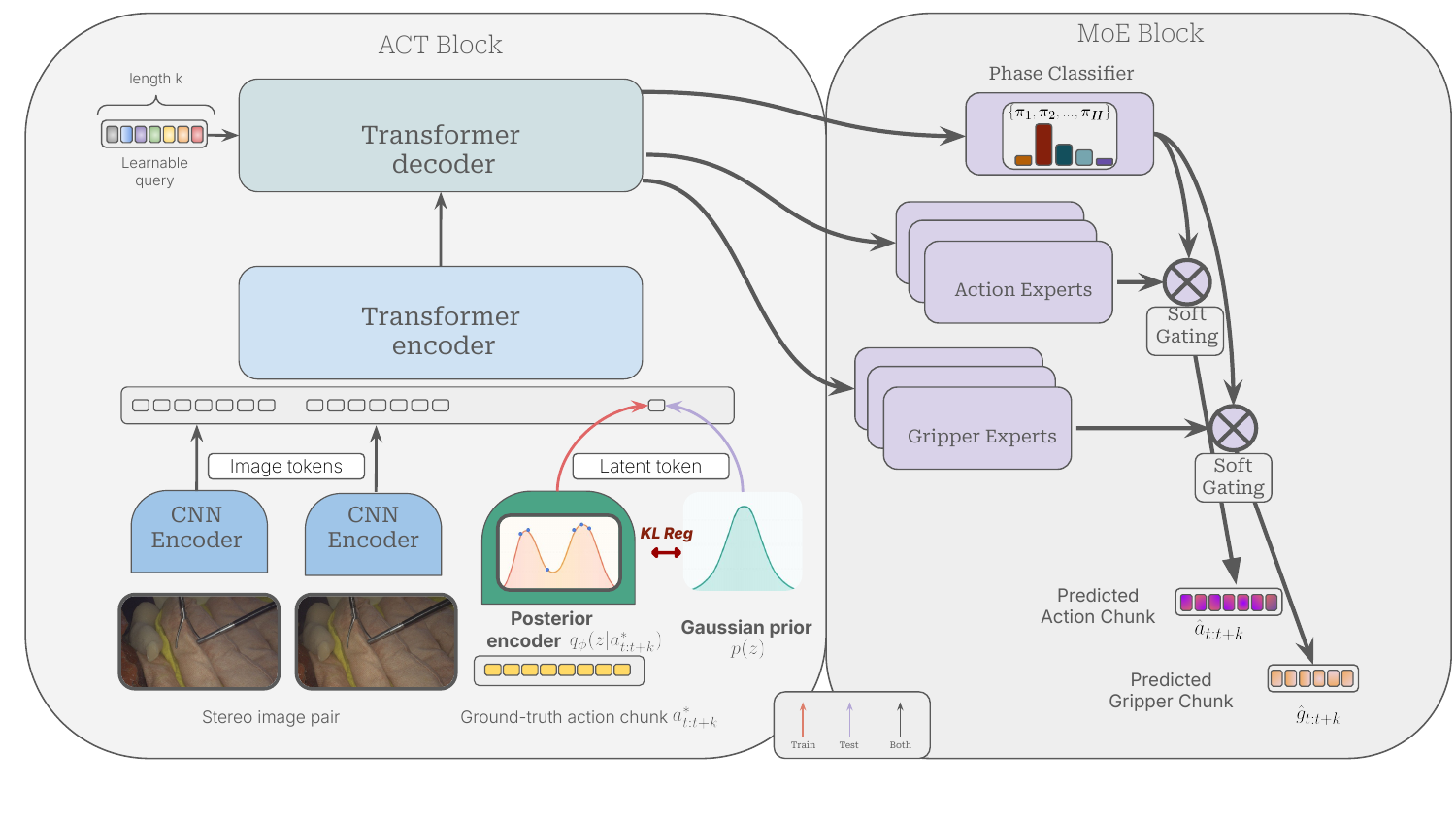}%
    \caption{Overview of the supervised Mixture-of-Experts extension to ACT. Stereo endoscopic images are processed by the ACT backbone to produce latent visual features and action tokens. A phase-aware gating network, supervised using surgical phase labels during training, routes features to specialized experts corresponding to different task phases (idle, grasping, holding, retraction, and tension maintenance). The outputs of the action and gripper experts are combined through phase-weighted aggregation to generate the action chunk.}
    \label{fig_model}
\end{figure*}

We record a total of 120 episodes with a fixed viewpoint of the scene from the endoscope. We refer to this dataset as the \textit{fixed-viewpoint dataset}.
During each trial, we record stereo image pairs from the endoscope, the binary state of the mechatronic interface (gripper open/closed), and the three-dimensional position of the instrument tip in the camera coordinate system.
We segment the bowel retraction task into $H=5$ phases as shown in table \ref{tab_phases}, with transitions primarily triggered by the surgeon's actions. The phases in the dataset are automatically labeled using the gripper state and the magnitude of movement of the robot instrument as a proxy for the transitions.
To improve the robustness of downstream policies, we introduce variability across the training demonstrations, including different starting points for the assistant tool position, different grasping locations indicated by the surgeon and slight movements of the bowel across the phantom scene. This ensures a diverse set of trajectories and visual features in the dataset.
To evaluate generalization to viewpoint variation, we additionally collect 50 episodes with randomized endoscopic camera angles. We refer to this dataset as the \textit{random-viewpoint dataset}.
\subsection{Observation and Action Space}
At time $t$, the observation space of the policy consists of the state $s_t = (I_t^{\text{left}}, I_t^{\text{right}})$ containing \textbf{only the stereo endoscopic image pair at the current time}. Although available, we explicitly exclude proprioceptive data to derive a vision-only policy. This design decouples performance from kinematic sensor noise and calibration drift, and makes it independent of the underlying surgical hardware quality \cite{kim2024surgical}.
The action space, similarly to \cite{act}, comprises chunks of $k$ continuous actions $a_{t:t+k}$, where each $a_{t+i} \in \mathbb{R}^3$ represents the delta movement of the instrument tip in the Cartesian space of the camera coordinate system, plus chunks of binary gripper actions $g_{t:t+k}$, where each $g_{t+i} \in \{0, 1\}$. We additionally denote ground truth phase label $h\in H$ at time $t$ with $\psi_t=h$. For action notations, we indicate ground truth using symbol $a^*$ and predictions using $\hat{a}$.
\subsection{Policy}\label{subsec_modelarch}
We propose a supervised Mixture-of-Experts (MoE) architecture that is modular and can be integrated into any transformer-based action chunking policy to leverage explicit task phase structure. Here, we apply it to the lightweight Action Chunking Transformer (ACT) \cite{act}, chosen for its efficiency, low latency, and strong performance on data-limited surgical tasks \cite{haworth2025suturebot}.
The base follows ACT’s variational framework: at training time, given a state observation $s_t$ and a chunk of size $k$ of ground-truth training actions $a^*_{t:t+k}$, a posterior encoder $q_\phi(z| a^*_{t:t+k})$ infers latent $z$, which is concatenated with visual features and processed through a transformer encoder-decoder with parameters $\theta$ to reconstruct action chunk $\hat{a}_{t:t+k}$. At inference, the posterior encoder is discarded and $z$ is set to the mean of a Gaussian prior $p(z) \sim \mathbb{N}(\mathbf{0},\mathbf{I})$.
We extend this with a Phase-Aware MoE block comprising $H$ parallel experts (one per phase), where each phase expert models the action distribution of a specific task phase, conditioned additionally on $z$ and $s_t$. 
The block is composed as following:
\begin{enumerate}
\item \textbf{Action Phase-Experts}: $H$ action heads, where expert $h$ outputs location parameters $\boldsymbol{\mu}_{h,t:t+k}(z, s_t, \psi = h) \in \mathbb{R}^{k \times d}$ for a $d$-dimensional action space over a chunk of length $k$. 
\item \textbf{Gripper Phase-Experts}: $H$ gripper heads, where expert $h$ outputs logits $\boldsymbol{\nu}_{h,t:t+k}(z, s_t, \psi = \psi_h) \in \mathbb{R}^{k}$,  parameterizing Bernoulli distributions with $p_h(g_{t+j}=1 \mid z, s_t, \psi = h) = \sigma(\nu_{h,t+j})$ for each $j^{th}$ gripper action of the chunk.
    \item \textbf{Gating Network}: A phase classifier that models the categorical distributions $\boldsymbol{\pi}_{t:t+k}(z, s_t)$ , where $\pi_{h,t+j} = p(\psi_{t+j}=h \mid z, s_t)$ and $\sum_{h=1}^H \pi_{h,t+j} = 1$.
\end{enumerate}

Final predictions are phase-weighted mixtures:
\begin{equation}
    \hat{a}_{t+j} = \sum_{h=1}^H \pi_{h,t+j} \cdot \mu_{h,t+j} \nonumber
\end{equation}
\begin{equation}
    \hat{g}_{t+j} = \sum_{h=1}^H \pi_{h,t+j} \cdot \sigma(\nu_{h,t+j}) \nonumber
\end{equation}
for each $(t+j)_{th}$ action of the chunk.
The architecture is illustrated in figure \ref{fig_model}. 
\subsection{Training Procedure}
Following the conditional variational framework of ACT \cite{act}, training optimizes a variational lower bound (ELBO) on the log-likelihood of demonstration trajectories \cite{ermon}, where the phase labels $\psi^*_{t:t+k}$ are observed during training. This supervised approach leverages privileged phase information to guide expert specialization during training.
By assuming Laplace-distributed action errors, Bernoulli gripper states, and categorical phase distributions, we obtain the following training objective \eqref{loss}, which comprises four components: i) the action reconstruction loss (L1) trains the weighted mixture of experts to match demonstration actions and provides robustness against outliers in human demonstrations \cite{act}; ii) the phase cross-entropy loss (CE) directly supervises the gating network, treating phase prediction as an auxiliary task, guiding the MoE experts to specialize per task phase; iii) the gripper binary cross-entropy loss (BCE) trains the weighted Bernoulli mixture for discrete gripper actions; iv) the KL term regularizes the learned amortized posterior encoder \cite{kingma2013auto, blei2017variational, higgins2017beta}, that we denote as $q_\phi$.
\begin{align}
\mathcal{L}(\theta, \phi) = &\alpha \sum_{j=0}^{k-1} \left\|\hat{a}_{t+j} - a^*_{t+j}\right\|_1 + \gamma \sum_{j=0}^{k-1} \text{CE}(\boldsymbol{\pi}_{t+j}, \psi^*_{t+j}) \nonumber \\
&+ \delta \sum_{j=0}^{k-1} \text{BCE}(\hat{g}_{t+j}, g^*_{t+j}) + \beta \cdot D_{\text{KL}}(q_\phi \| p(z))
\label{loss}
\end{align}
As benchmarks, we train ACT, SmolVLA and $\pi_{0.5}$ with standard hyperparameters, using the open-source LeRobot codebase \cite{anonymous2025lerobot}. As SmolVLA and $\pi_{0.5}$ include language instruction and proprioceptive state in their input space, we extend $s_t$ with a fixed language instruction and a padding proprioceptive vector of zeros. Table \ref{tab:model_params} summarizes the evaluated models, while additional implementation details are provided in the supplementary material.
We train and deploy each policy on a single RTXA5000 Nvidia GPU, with the exception of $\pi_{0.5}$ which required one A100 Nvidia GPU for training. The training time of SmolVLA was 14 hours, 8 hours for $\pi_{0.5}$ and 3 hours for ACT and ACT+MoE.

\begin{table}[h]
    \centering
    \caption{Comparison of model variants, parameter counts, and training frameworks.}
    \label{tab:model_params}
    \begin{tabular}{llcl}
        \toprule
        \textbf{Policy} & \textbf{Variant} & \textbf{Parameters} & \textbf{Codebase} \\
        \midrule
        $\pi_{0.5}$ & Base & 4 B & LeRobot \\
        SmolVLA & Small & 0.24 B & LeRobot \\
        ACT+MoE(Ours) & -- & 53.3 M & Ours \\
        ACT & -- & 52 M & Ours \\
        \bottomrule
    \end{tabular}
\end{table}
\section{Experiments}\label{sec_experiments}
To evaluate the success rate of the learned policies on the robotic platform, two trained medical students and one surgical resident review each policy roll-out in a single-blinded process and label the final frame as either a success or a failure. A success in the final frame is defined as tissue grasped by two graspers and retracted with sufficient tension on the bowel segment. The final outcome of each roll-out is then determined by majority voting.
We first evaluate the policy trained on the fixed viewpoint dataset using environment conditions as close as possible to the training scene and the same camera angle. We denote these tests as \textit{in-distribution}.
Consequently, we assess generalization capabilities by testing four out-of-distribution conditions: i) grasping bowel sections not seen during training, ii) operating under severely reduced scene illumination, iii) handling partial occlusions from phantom fat, and iv) using a slightly different camera angle compared to the training data. We denote these policy roll-outs as  \textit{out-of-distribution}.
For the in-distribution evaluation, the endoscopic field of view was divided into 12 angular sectors. The robotic instrument was initialized from 20 distinct starting positions distributed across these sectors. Each starting position was paired with one of eight distinct grasping points on the bowel, resulting in 20 unique starting position and grasping point combinations per policy. For the out-of-distribution evaluation, we performed five trials for each perturbation type, sampling starting positions from the same 20-position set used in-distribution. The novel-grasp-location perturbation differed only in the target grasp locations: the surgeon indicated targets on a third bowel segment at image coordinates not observed during training. This setting directly tests whether the policy follows the surgeon's visual cue rather than memorizing a fixed spatial prior.
The best-performing policy (based on success rate) is tested zero-shot on \textit{ex vivo} porcine bowel, replacing the phantom bowel in the experimental setup. We conduct 15 trials in the \textit{ex vivo} configuration to evaluate whether the learned policies transfer to real tissue without additional training.
Additionally we train again the best performing policy on the random viewpoint dataset, and test it on unseen camera angles, to verify its robustness to the 3D scene in prevision of \textit{in vivo} conditions. 

\section{Results and Discussion}\label{sec_results}
\subsection{In-Distribution Roll-Outs} 
We report the success rates of the policies for in-distribution experiments in table \ref{table_success_rate}.
\begin{table*}[h]
\centering
\begin{tabular*}{\textwidth}{@{\extracolsep\fill}lccccc@{}}
\toprule
Policy & \multicolumn{5}{c}{In-Distribution} \\ 
\cmidrule(lr){2-6}
& Reaching & Grasping & Retracting & End-To-End  & Inference Frequency\\
\midrule
ACT  & 16/20 & 13/20 & 12/20 & 10/20  & 27 Hz \\
$\pi_{0.5}$   & 2/20 & 4/20 & 0/20 & 0/20  & 10 Hz (compiled)\\
SmolVLA  & 5/20 & 2/20 & 0/20 & 0/20  & 3.3 Hz\ \\
ACT + MoE (Ours)     & \textbf{20/20}     & \textbf{20/20}$^{*}$ & \textbf{19/20}$^{*}$ & \textbf{17/20}$^{*}$ & 27 Hz \\\bottomrule
\end{tabular*}
\caption{Success rates (divided per sub-task) of policies trained on the fixed viewpoint dataset. In-distribution roll-outs inside the phantom environment.}
\label{table_success_rate}
\begin{flushleft} \footnotesize
\textit{$^{*}$ indicates statistically significant improvement of ACT + MoE over the ACT baseline (two-sided Fisher's exact test, $p < 0.05$; exact $p$-values: Grasping $p = 0.008$, Retracting $p = 0.020$, End-To-End $p = 0.041$). Improvements of ACT + MoE over both $\pi_{0.5}$ and SmolVLA are highly significant ($p < 10^{-7}$) on all reported metrics.}
\end{flushleft}
\end{table*}

Our benchmarking reveals a significant performance gap between generalist VLA models and specialized action transformers. Both VLA baselines fail to complete the task end-to-end. SmolVLA proves unable to model the trajectory dynamics, producing erratic and dangerous actions against the target anatomy. While $\pi_{0.5}$ demonstrates a slight improvement in grasping capabilities, it suffers from severe temporal incoherence; the model frequently violates task phase constraints, initiating retraction motions before securing the grasp or anticipating the surgeon's handover prematurely. Consequently, it achieves a 0\% end-to-end success rate. We argue that both models fail to model the task due to the limited amount of training data available compared to the actual number of parameters that they need to optimize for, shown in Table \ref{tab:model_params}.

In contrast, the standard ACT baseline demonstrates reasonable competency, achieving a 50\% success rate. However, it lacks fine-grained dexterity, frequently resulting in tissue slippage or imprecise end-effector positioning during critical phase transitions.
Incorporating our Supervised MoE architecture yields a significant performance gain. The specialized experts enable precise phase handling, boosting the grasping success rate from 60\% (ACT) to 85\% and the overall end-to-end success rate from 50\% to 85\%. This represents a \textit{70\% relative improvement} over the standard ACT baseline, confirming that explicit expert supervision significantly enhances policy robustness and dexterity in data-constrained regimes.  In terms of computational efficiency, our approach maintains the real-time applicability of the base architecture.  ACT + MoE policy operates at 27 Hz, incurring negligible inference overhead compared to standard ACT. Conversely, VLA baselines exhibit significantly higher latency—with $\pi_{0.5}$ running at 10 Hz and SmolVLA at 3.3 Hz—rendering them impractical for the high-frequency control loops required in surgical automation. 
\subsection{Out-of-Distribution Roll-Outs}
Given the complete failure of VLA models in the standard setting, we exclude them from further evaluation. We restrict the out-of-distribution (OOD) analysis to a comparison between the standard ACT baseline and our MoE-augmented policy. The results are detailed in Table \ref{table_success_rate_ood}.
\begin{table}[h]
\centering
\begin{tabular*}{\columnwidth}{@{\extracolsep\fill}lcccc@{}}
\toprule
Policy & \multicolumn{4}{c}{Out-of-Distribution} \\ 
\cmidrule(lr){2-5}
& Reaching & Grasping & Retracting & End-To-End\\
\midrule
ACT  & 16/20 & 13/20 & 6/20 & 6/20 \\
ACT + MoE (Ours)  & \textbf{19/20} & \textbf{16/20} & \textbf{13/20} & \textbf{13/20} \\
\bottomrule
\end{tabular*}
\caption{Success rates (divided per sub-task) of policies trained on the fixed viewpoint dataset. OOD roll-outs inside the phantom environment.}
\label{table_success_rate_ood}
\begin{flushleft} \footnotesize
\textit{two-sided Fisher's exact test $p$-values: Reaching $p = 0.342$, Grasping $p =0.480$, Retracting $p = 0.056$, End-To-End $p = 0.056$.
}
\end{flushleft}

\end{table}
\subsection{Ex Vivo Zero-Shot Roll-Outs} 
Motivated by the superior out-of-distribution performance of our supervised MoE-ACT, we select it for the next two tests: i) zero-shot testing on \textit{ex vivo} porcine bowel and ii) retraining on the random-viewpoint dataset and testing on unseen viewpoints. The model achieves an \textbf{80\%} success rate on \textit{ex vivo} porcine bowel (\textbf{12/15}) (Figure \ref{fig_frontcover}, central panel), validating its generalization capabilities. Of the three failures, two were due to grasping two bowels at the same time — but still completing the retraction — and only one consisted of a complete failure.
\subsection{Random Viewpoint Roll-Outs} 

\begin{figure}[t]
    \includegraphics[width=\columnwidth]{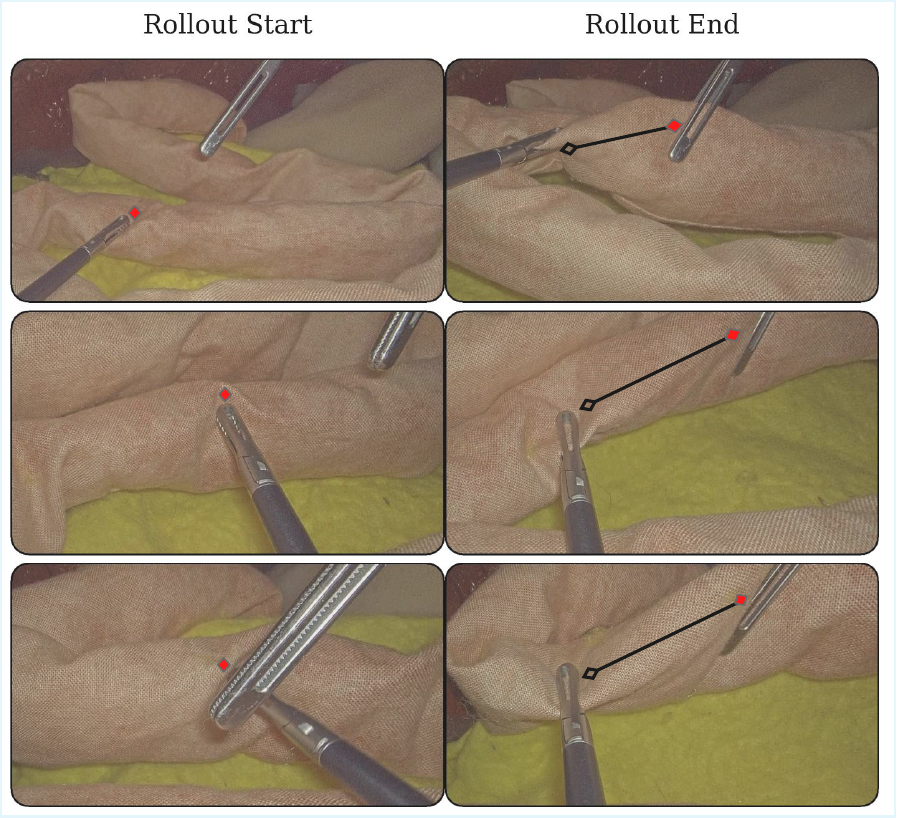}%
    \caption{Roll-outs of our policy trained on the random-viewpoint dataset generalize on unseen camera viewpoints, showing robust performance across zoom and orientation changes. Examples show initial (left) and final (right) frames of the roll-outs. Human-controlled surgeon instrument on the bottom, robot-controlled assistant instrument on the top.}
    \label{fig_multi_cam_examples}
\end{figure}

We further evaluate the ability of MoE-ACT to generalize to diverse camera viewpoints. We train the policy on the joint combination of the fixed and random-viewpoint datasets, to simulate the more realistic viewing conditions of clinical \textit{in vivo} procedures, where precise camera positioning cannot be fixed or known a priori.

After retraining, our policy roll-outs achieve an 82\% success rate (18/22) on unseen testing viewpoints, demonstrating robust performance and implicit 3D scene understanding. In figure \ref{fig_multi_cam_examples} we show representative examples of the diverse set of camera angles used during testing, with considerable variations in both zoom and orientation levels. These variations more closely approximate the random viewing conditions that autonomous policies should be robust against when translated to real surgical scenarios.

\subsection{Qualitative Analysis and Ablation Studies}
\begin{figure}[t]
    \includegraphics[width=\columnwidth]{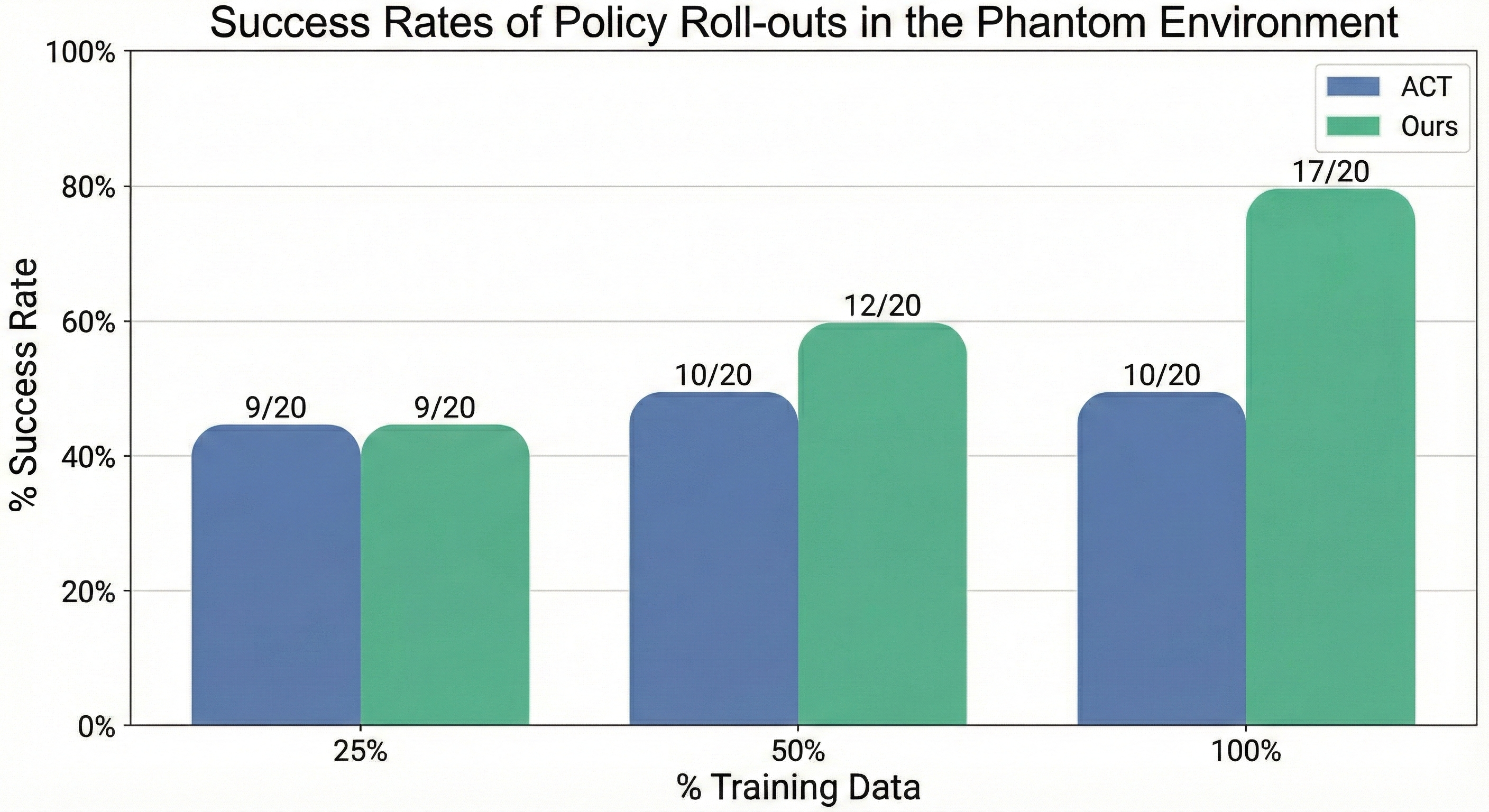}%
    \caption{Ablation on the amount of training data demonstrations plotted against the policy success rates for In-Distribution Roll-Outs}
    \label{fig_scaling_ablation}
\end{figure}
\begin{figure}[t]
    \includegraphics[width=\columnwidth]{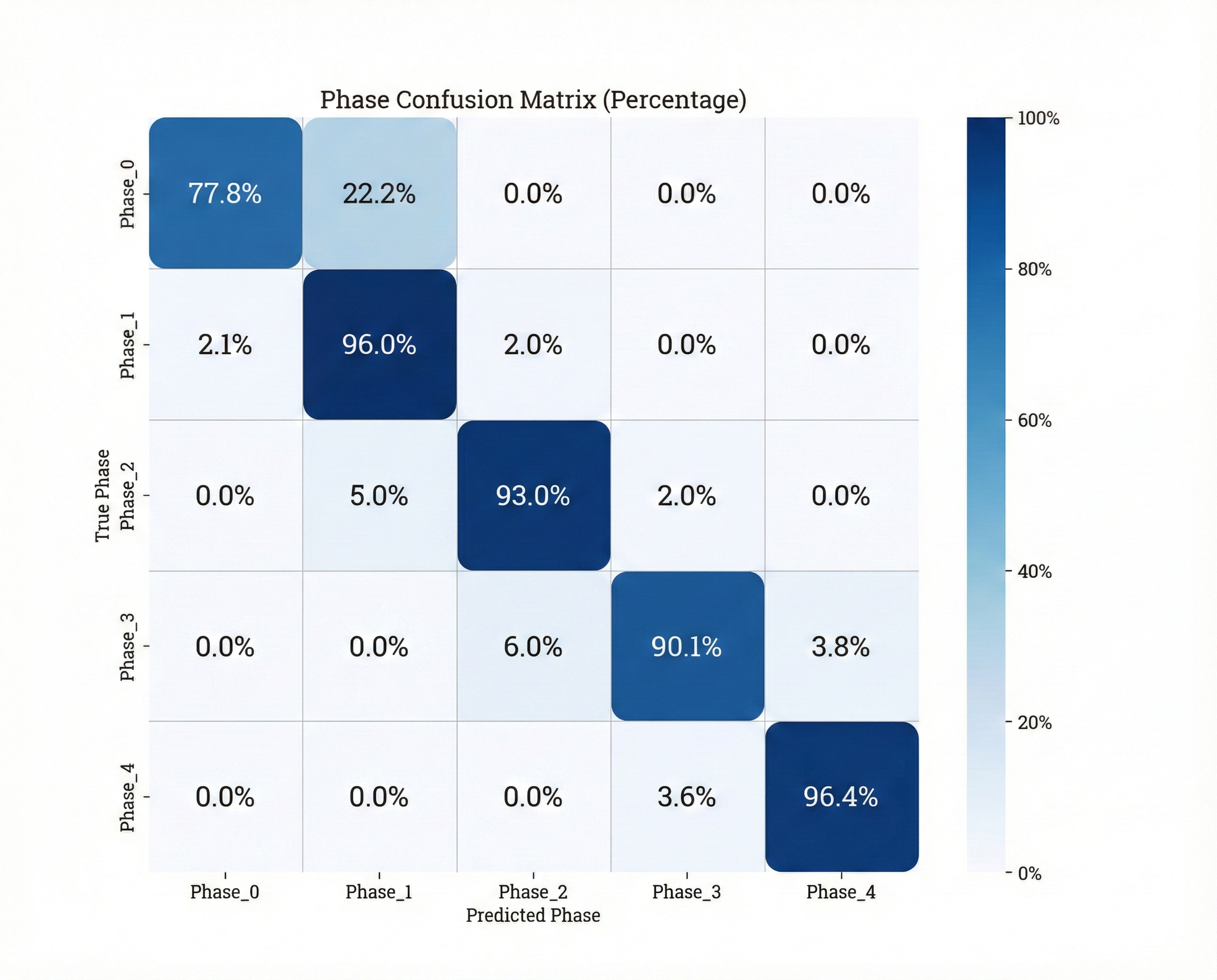}%
   \caption{Confusion matrix of the MoE gating network on the validation dataset, demonstrating the effectiveness of the auxiliary phase classification task.}
    \label{fig_confmatrix}
\end{figure}
\begin{figure}[t]
    \includegraphics[width=0.8\columnwidth]{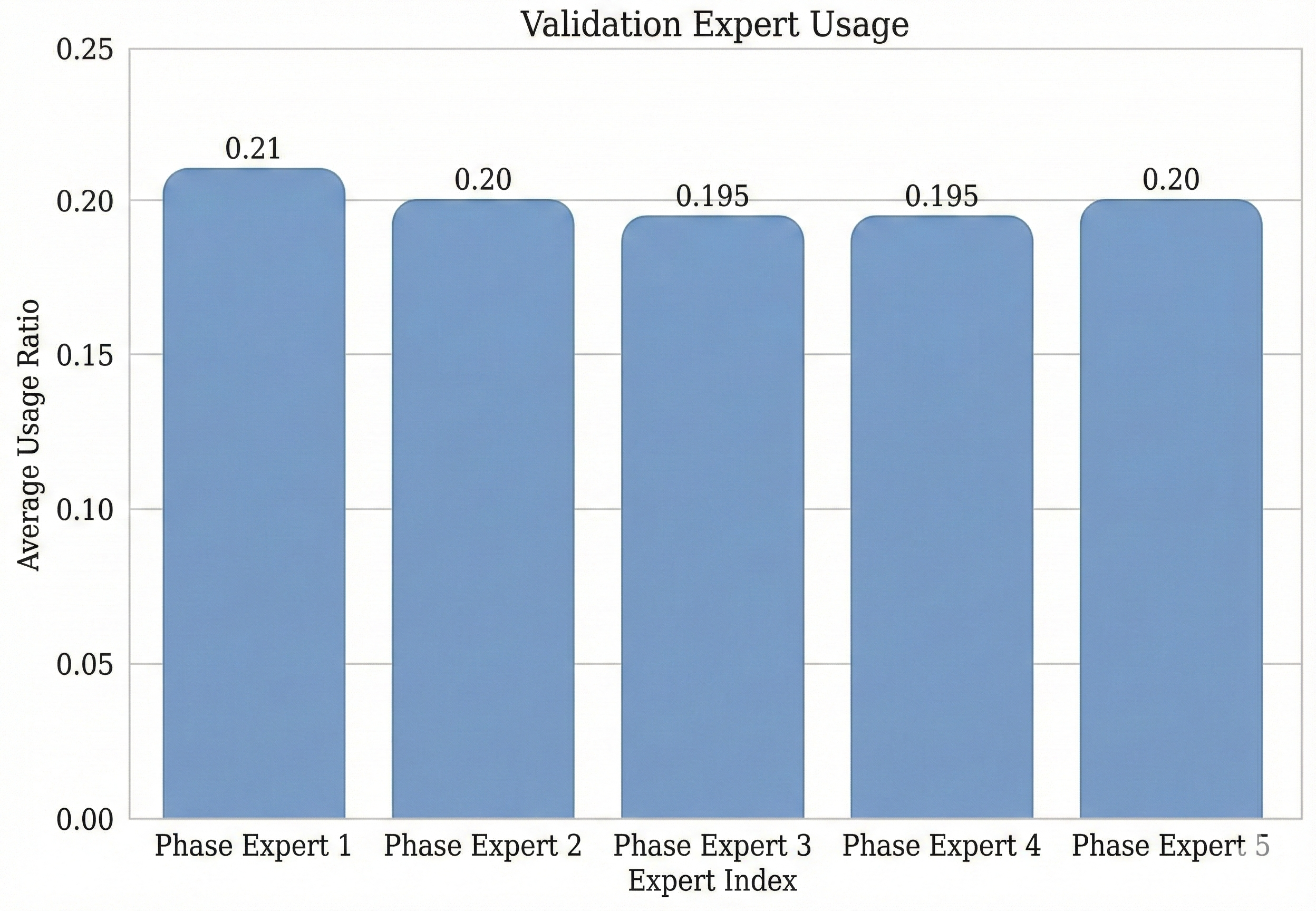}%
   \caption{Expert utilization rates on the validation dataset. The activation frequency aligns with the distribution of task phases in the training data.}
    \label{fig_expusage}
\end{figure}
\begin{figure}[htbp]
\centering
\begin{minipage}[b]{0.45\textwidth}
    \includegraphics[width=\textwidth, trim=0 100 0 50, clip]{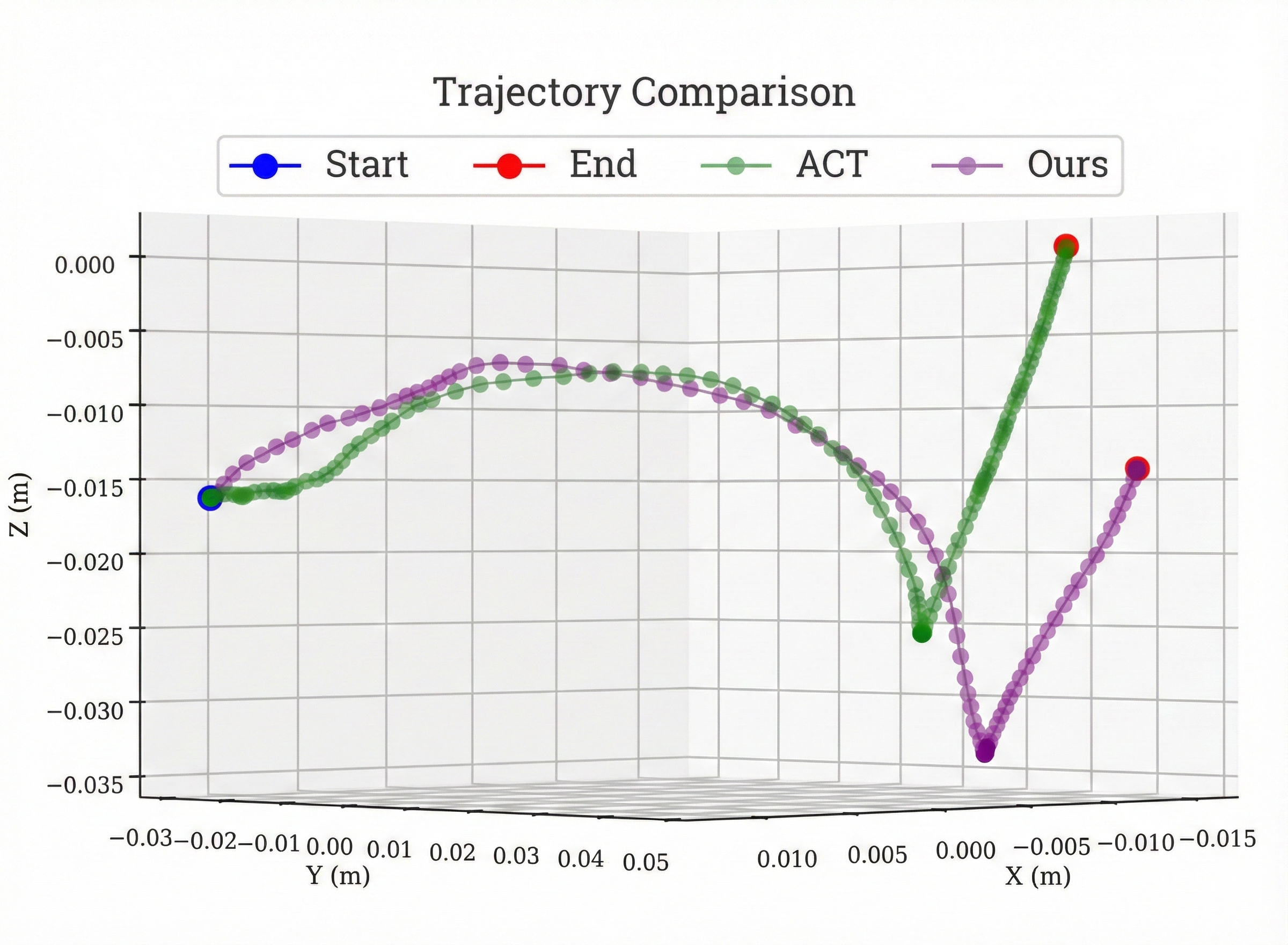} \\[1mm]
\end{minipage}
\hspace{2mm}
\begin{minipage}[b]{0.30\textwidth}
    \includegraphics[width=\textwidth]{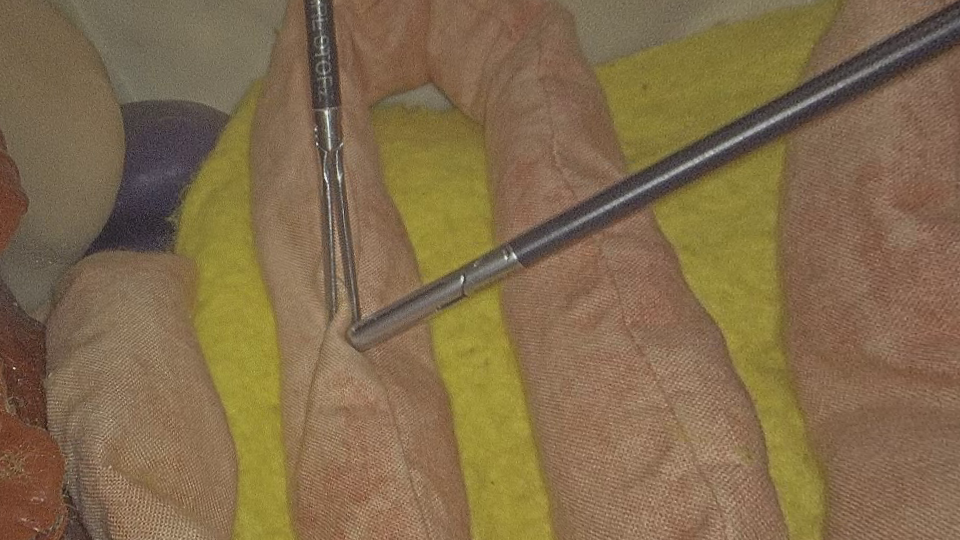} \\[2mm]  
    \includegraphics[width=\textwidth]{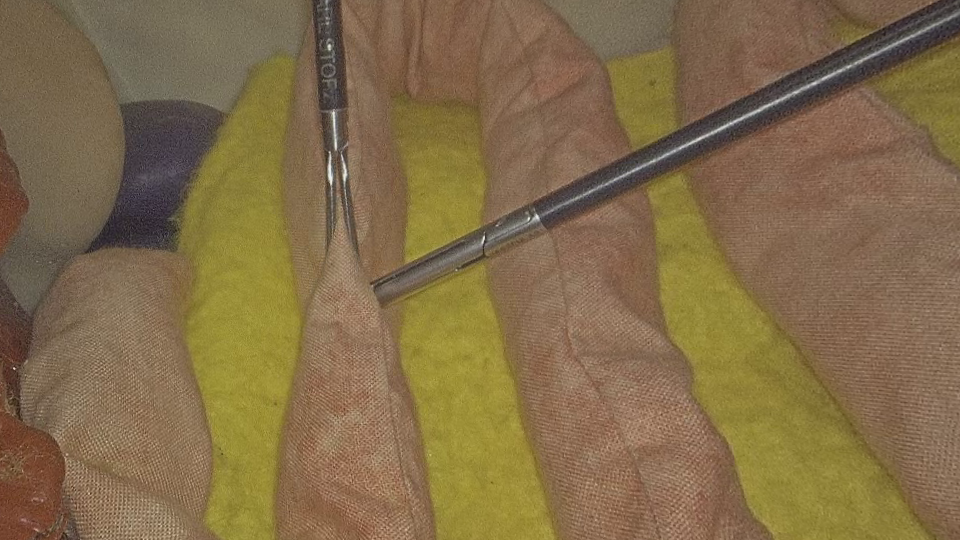}
\end{minipage}

\caption{Top: trajectories of ACT and our ACT + MoE decoder from the same starting position to the same target grasping point, showing approach and retraction of the instrument. Bottom: frames of ACT (top) and ACT + MoE (bottom) highlight the difference in grasping depth. The MoE policy visibly demonstrates a deeper and more secure grasp, a consistent pattern observed during policy roll-outs. Robot-controlled instrument is grasping, human-controlled instrument is pointing.}
\label{fig_trajectory}
\end{figure}

\begin{figure*}
\centering
\begin{tabular}{cc}
    \includegraphics[width=0.3\textwidth]{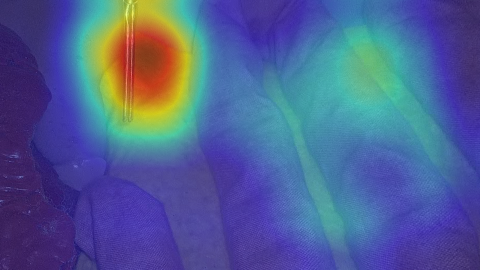} &
    \includegraphics[width=0.3\textwidth]{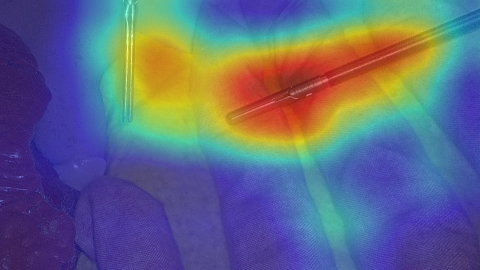} \\
    \includegraphics[width=0.3\textwidth]{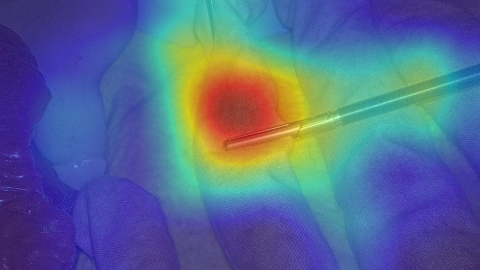} &
    \includegraphics[width=0.3\textwidth]{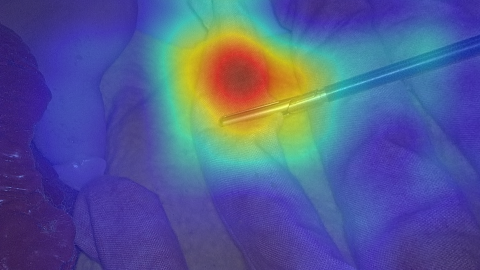}
\end{tabular}
\caption{AblationCAM heatmaps: the policy vision encoder focuses first on the robot instrument (top left), then on the surgeon instrument (top right), and at the end on stretching the bowel (bottom)}\label{heatmap}
\end{figure*}

 We first analyze the qualitative behavior of the policies to understand the performance gap observed in the main benchmarks. We find that the ACT baseline frequently exhibits superficial grasping behavior, characterized by insufficient tissue purchase. As illustrated in Fig. \ref{fig_trajectory}, the baseline's end-effector trajectories often fail to achieve the necessary approach depth compared to our MoE-augmented policy. This results in weak engagement with the phantom bowel tissue, leading to frequent slippage during the sustained retraction phase.
 
To evaluate the data efficiency of our approach, we performed an ablation study by training both the baseline and our MoE policy on progressively smaller subsets of the phantom dataset: 100\% (120 episodes), 50\% (60 episodes), and 25\% (30 episodes), with a consistent 10\% validation split. This analysis aims to investigate learning robustness under conditions of extreme data scarcity and identify the minimum dataset size required for learning multi-step surgical tasks. 
The results, shown in Fig. \ref{fig_scaling_ablation}, show that in the lowest data regime (25\%), both policies perform identically (45\% success rate), suggesting that 30 demonstrations represent a lower bound where data scarcity bottlenecks performance regardless of architecture. However, as data availability increases, a significant divergence emerges: while the standard ACT baseline plateaus at 50\% success, our MoE policy effectively leverages the additional data, scaling from 60\% to 85\% success.

We further extend our analysis by examining the learned feature representations of the vision encoder of our MoE policy during the inference roll-outs. We construct saliency maps using an adaptation of AblationCAM \cite{desai2020ablation} for regression tasks, where we ablate regions of the visual feature maps and measure the resulting change in the predicted action norm, identifying which visual regions most strongly influence the policy's action magnitude.
As shown in Fig. \ref{heatmap}, the resulting saliency maps reveal distinct patterns across task phases: early in episodes, high saliency concentrates on the robotic instrument's position; when the surgeon provides visual cues, saliency shifts to the indicated grasping target; during retraction, the policy exhibits high sensitivity to the bowel segment between the robot's and surgeon's grasping points. This phase-dependent shift in salient regions suggests that the policy has learned to extract task-relevant visual features for each phase exploiting successfully its MoE block.

We analyze the performance of the phase classifier serving as the gating network for the action experts. The validation set confusion matrix, presented in Fig. \ref{fig_confmatrix}, demonstrates high classification accuracy, confirming that surgical phase classification is effectively learned as an auxiliary task. This explicit supervision enables the policy to correctly route states to their specialized experts. Furthermore, we examine the expert utilization rates in Fig. \ref{fig_expusage}. We observe that the expert activation distribution closely mirrors the phase frequency of the training dataset. This validates that our method ensures balanced expert specialization, effectively preventing the policy from suffering from mode collapse or expert underutilization—common failure modes in unsupervised mixture-of-experts training.
\begin{figure}[t]
    \includegraphics[width=\columnwidth]{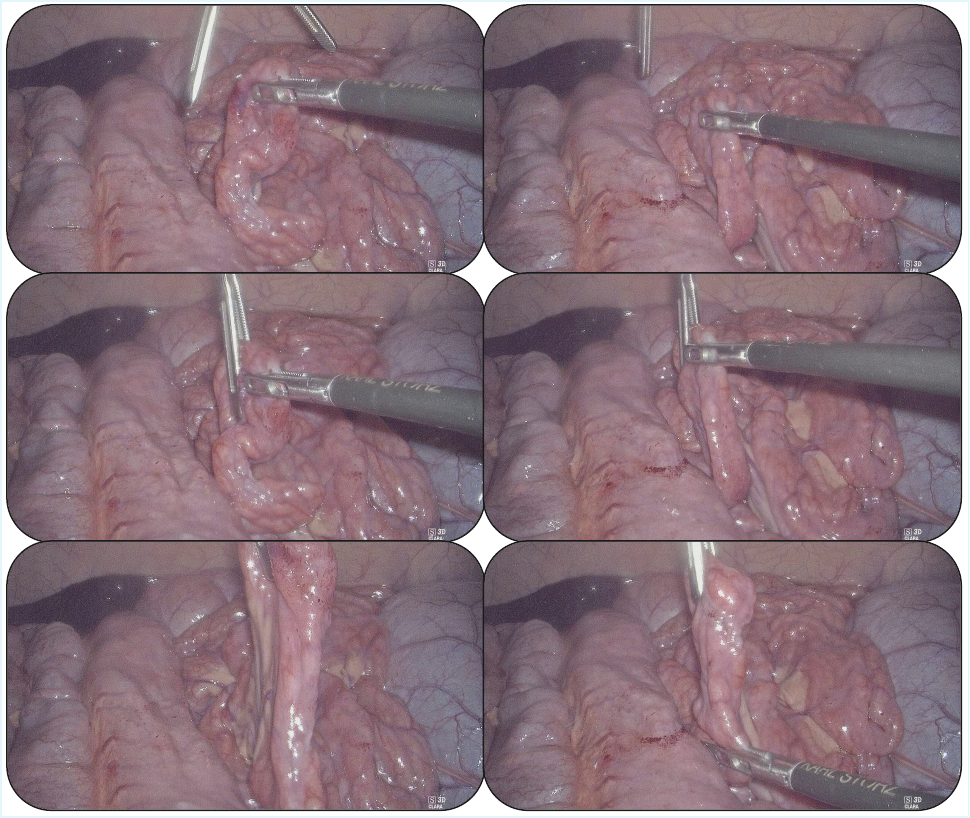}%
    \caption{Qualitative examples of two roll-outs of MoE-ACT policy during \textit{in vivo} porcine surgery. Rows show task phases: reach (top), grasp (middle) and retract (bottom). Robot-controlled instrument on the top, human-controlled instrument on the right.}
    \label{fig_invivo}
\end{figure}

Finally, we show preliminary qualitative results of the supervised MoE policy during \textit{in vivo} porcine surgery\footnote{\tiny All procedures were approved by the local state authority (TVV43/2023, Saxony, Germany), and conducted in accordance with institutional ethical standards for animal experimentation and the registered protocol.}. In Fig. \ref{fig_invivo}, we show two examples of successful policy rollouts in the aforementioned setting. We plan to extensively evaluate and present quantitative results in an \textit{in vivo} porcine environment in our future work.

\section{Conclusion}\label{sec_conclusion}
In this work, we presented a Supervised Mixture-of-Experts architecture that enables lightweight action transformer policies to perform multi-step surgical manipulation tasks from limited data. 
Additionally, we demonstrated that a purely vision-based policy can generalize to unseen camera viewpoints if trained with sufficient geometric variability. Our experiments show that by incorporating random viewpoints during training, the policy maintains high success rates on unseen camera angles at test time, effectively preventing overfitting to a static 3D scene configuration. This capability, combined with successful zero-shot transfer to \textit{ex vivo} tissue, underscores the potential of our method for real-world surgical assistance where the endoscopic view is not known a priori.

Limitations of our current framework include the reliance on manual phase supervision to guide the MoE gating network. Future work will investigate unsupervised learning methods to discover latent task skills implicitly from demonstration data. Moreover, we plan to equip our policies with real-time depth vision derived from stereoscopic endoscope feed to further enhance 3D spatial understanding and robustness for \textit{in vivo} deployment. This is motivated by the observed behaviour of the policy in the preliminary \textit{in vivo} experiments, where the lack of depth is an evident factor for the failure of the task in the grasping phase. We include examples of this in the supplementary material. 

In conclusion, this work has shown that Imitation Learning coupled with a Mixture-of-Experts architecture can automate multi-step minimally-invasive surgical tasks using only endoscopic vision, building toward the goal of deploying learned policies \textit{in vivo}.

\section*{Acknowledgments}
The authors would like to acknowledge the support of the Federal Ministry of Research, Technology, and Space (BMFTR) within the research program “Communication Systems: Souverän. Digital. Vernetzt.”, Joint Project 6G-life (Project ID: 16KIS2413K). This work is also supported by the project “Next Generation AI Computing (gAIn)”, funded by the Bavarian Ministry of Science and the Arts and the Saxon Ministry for Science, Culture, and Tourism. The authors further acknowledge support from the German Research Foundation (DFG, Deutsche Forschungsgemeinschaft) under the Germany's Excellence Strategy - EXC 2050/2 - Project ID 390696704 - Cluster of Excellence "Centre for Tactile Internet with Human-in-the-Loop" (CeTI) at TUD Dresden University of Technology. In addition, this work has been partially supported by the Federal Ministry of Research, Technology, and Space (BMFTR) within the Robotics Institute Germany (RIG), grant agreement no. 16ME1005.

\clearpage


\bibliographystyle{plainnat}
\bibliography{references}

@article{chi2023diffusion,
  title={\href{https://www.roboticsproceedings.org/rss19/p026.pdf}{Diffusion policy: Visuomotor policy learning via action diffusion}},
  author={Chi, Cheng and Xu, Zhenjia and Feng, Siyuan and Cousineau, Eric and Du, Yilun and Burchfiel, Benjamin and Tedrake, Russ and Song, Shuran},
  journal={The International Journal of Robotics Research},
  pages={02783649241273668},
  year={2023},
  publisher={SAGE Publications Sage UK: London, England}
}

@InProceedings{pmlr-v305-black25a,
  title = 	 {\href{https://www.pi.website/download/pi05.pdf}{$\pi_{0.5}$: a Vision-Language-Action Model with Open-World Generalization}},
  author =       {Black, Kevin and Brown, Noah and Darpinian, James and Dhabalia, Karan and Driess, Danny and Esmail, Adnan and Equi, Michael Robert and Finn, Chelsea and Fusai, Niccolo and Galliker, Manuel Y. and Ghosh, Dibya and Groom, Lachy and Hausman, Karol and ichter, brian and Jakubczak, Szymon and Jones, Tim and Ke, Liyiming and LeBlanc, Devin and Levine, Sergey and Li-Bell, Adrian and Mothukuri, Mohith and Nair, Suraj and Pertsch, Karl and Ren, Allen Z. and Shi, Lucy Xiaoyang and Smith, Laura and Springenberg, Jost Tobias and Stachowicz, Kyle and Tanner, James and Vuong, Quan and Walke, Homer and Walling, Anna and Wang, Haohuan and Yu, Lili and Zhilinsky, Ury},
  booktitle = 	 {Proceedings of The 9th Conference on Robot Learning},
  pages = 	 {17--40},
  year = 	 {2025},
  editor = 	 {Lim, Joseph and Song, Shuran and Park, Hae-Won},
  volume = 	 {305},
  series = 	 {Proceedings of Machine Learning Research},
  month = 	 {27--30 Sep},
  publisher =    {PMLR},
  url = 	 {https://proceedings.mlr.press/v305/black25a.html}
}

@article{shukor2025smolvla,
  title={\href{https://arxiv.org/pdf/2506.01844}{Smolvla: A vision-language-action model for affordable and efficient robotics}},
  author={Shukor, Mustafa and Aubakirova, Dana and Capuano, Francesco and Kooijmans, Pepijn and Palma, Steven and Zouitine, Adil and Aractingi, Michel and Pascal, Caroline and Russi, Martino and Marafioti, Andres and others},
  journal={arXiv preprint arXiv:2506.01844},
  year={2025}
}

@article{pertsch2025fast,
  title={\href{https://www.pi.website/download/fast.pdf}{Fast: Efficient action tokenization for vision-language-action models}},
  author={Pertsch, Karl and Stachowicz, Kyle and Ichter, Brian and Driess, Danny and Nair, Suraj and Vuong, Quan and Mees, Oier and Finn, Chelsea and Levine, Sergey},
  journal={arXiv preprint arXiv:2501.09747},
  year={2025}
}

@article{zhou2024variational,
  title={\href{https://arxiv.org/pdf/2406.12538}{Variational distillation of diffusion policies into mixture of experts}},
  author={Zhou, Hongyi and Blessing, Denis and Li, Ge and Celik, Onur and Jia, Xiaogang and Neumann, Gerhard and Lioutikov, Rudolf},
  journal={Advances in Neural Information Processing Systems},
  volume={37},
  pages={12739--12766},
  year={2024}
}

@inproceedings{
anonymous2025lerobot,
title={\href{https://openreview.net/pdf?id=CiZMMAFQR3}{LeRobot:  An Open-Source Library for End-to-End Robot Learning}},
author={Anonymous},
booktitle={Submitted to The Fourteenth International Conference on Learning Representations},
year={2025},
url={https://openreview.net/forum?id=CiZMMAFQR3},
note={under review}
}

@article{khazatsky2024droid,
  title={\href{https://arxiv.org/html/2403.12945v1}{Droid: A large-scale in-the-wild robot manipulation dataset}},
  author={Khazatsky, Alexander and Pertsch, Karl and Nair, Suraj and Balakrishna, Ashwin and Dasari, Sudeep and Karamcheti, Siddharth and Nasiriany, Soroush and Srirama, Mohan Kumar and Chen, Lawrence Yunliang and Ellis, Kirsty and others},
  journal={arXiv preprint arXiv:2403.12945},
  year={2024}
}

@article{kim2025fine,
  title={\href{https://arxiv.org/pdf/2502.19645}{Fine-tuning vision-language-action models: Optimizing speed and success}},
  author={Kim, Moo Jin and Finn, Chelsea and Liang, Percy},
  journal={arXiv preprint arXiv:2502.19645},
  year={2025}
}

@INPROCEEDINGS{act, 
    AUTHOR    = {Tony Z. Zhao AND Vikash Kumar AND Sergey Levine AND Chelsea Finn}, 
    TITLE     = {\href{https://roboticsproceedings.org/rss19/p016.pdf}{Learning Fine-Grained Bimanual Manipulation with Low-Cost Hardware}}, 
    BOOKTITLE = {Proceedings of Robotics: Science and Systems}, 
    YEAR      = {2023}, 
    ADDRESS   = {Daegu, Republic of Korea}, 
    MONTH     = {July}, 
    DOI       = {10.15607/RSS.2023.XIX.016} 
}

@article{jacobs1991adaptive,
  title={\href{https://www.cs.toronto.edu/~fritz/absps/jjnh91.pdf}{Adaptive mixtures of local experts}},
  author={Jacobs, Robert A and Jordan, Michael I and Nowlan, Steven J and Hinton, Geoffrey E},
  journal={Neural computation},
  volume={3},
  number={1},
  pages={79--87},
  year={1991},
  publisher={MIT Press}
}

@article{jordan1994hierarchical,
  title={\href{https://www.cs.toronto.edu/~hinton/absps/hme.pdf}{Hierarchical mixtures of experts and the EM algorithm}},
  author={Jordan, Michael I and Jacobs, Robert A},
  journal={Neural computation},
  volume={6},
  number={2},
  pages={181--214},
  year={1994},
  publisher={MIT Press}
}

@article{blei2017variational,
  title={\href{https://www.cs.columbia.edu/~blei/fogm/2018F/materials/BleiKucukelbirMcAuliffe2017.pdf}{Variational inference: A review for statisticians}},
  author={Blei, David M and Kucukelbir, Alp and McAuliffe, Jon D},
  journal={Journal of the American Statistical Association},
  year={2017}
}

@inproceedings{ermon,
author = {Yu, Lantao and Yu, Tianhe and Song, Jiaming and Neiswanger, Willie and Ermon, Stefano},
title = {\href{https://doi.org/10.1609/aaai.v37i9.26305}{Offline imitation learning with suboptimal demonstrations via relaxed distribution matching}},
year = {2023},
isbn = {978-1-57735-880-0},
publisher = {AAAI Press},
url = {https://doi.org/10.1609/aaai.v37i9.26305},
doi = {10.1609/aaai.v37i9.26305},
booktitle = {Proceedings of the Thirty-Seventh AAAI Conference on Artificial Intelligence and Thirty-Fifth Conference on Innovative Applications of Artificial Intelligence and Thirteenth Symposium on Educational Advances in Artificial Intelligence},
articleno = {1236},
numpages = {9},
series = {AAAI'23/IAAI'23/EAAI'23}
}

@inproceedings{higgins2017beta,
  title={\href{https://www.cs.toronto.edu/~bonner/courses/2022s/csc2547/papers/generative/disentangled-representations/beta-vae,-higgins,-iclr2017.pdf}{beta-VAE: Learning basic visual concepts with a constrained variational framework}},
  author={Higgins, Irina and others},
  booktitle={ICLR},
  year={2017}
}

@article{kingma2013auto,
  title={\href{https://arxiv.org/pdf/1312.6114}{Auto-encoding variational bayes}},
  author={Kingma, Diederik P and Welling, Max},
  journal={arXiv preprint arXiv:1312.6114},
  year={2013}
}

@article{maier2017surgical,
  title={\href{https://www.nature.com/articles/s41551-017-0132-7}{Surgical data science for next-generation interventions}},
  author={Maier-Hein, Lena and Vedula, Swaroop S and Speidel, Stefanie and Navab, Nassir and Kikinis, Ron and Park, Adrian and Eisenmann, Matthias and Feussner, Hubertus and Forestier, Germain and Giannarou, Stamatia and others},
  journal={Nature Biomedical Engineering},
  volume={1},
  number={9},
  pages={691--696},
  year={2017},
  publisher={Nature Publishing Group UK London}
}

@inproceedings{funke2019using,
  title={\href{https://arxiv.org/pdf/1907.11454}{Using 3D convolutional neural networks to learn spatiotemporal features for automatic surgical gesture recognition in video}},
  author={Funke, Isabel and Bodenstedt, Sebastian and Oehme, Florian and von Bechtolsheim, Felix and Weitz, J{\"u}rgen and Speidel, Stefanie},
  booktitle={International conference on medical image computing and computer-assisted intervention},
  pages={467--475},
  year={2019},
  organization={Springer}
}

@article{kim2025srt,
  title={\href{https://arxiv.org/pdf/2505.10251}{SRT-H: A hierarchical framework for autonomous surgery via language-conditioned imitation learning}},
  author={Kim, Ji Woong and Chen, Juo-Tung and Hansen, Pascal and Shi, Lucy Xiaoyang and Goldenberg, Antony and Schmidgall, Samuel and Scheikl, Paul Maria and Deguet, Anton and White, Brandon M and Tsai, De Ru and others},
  journal={Science robotics},
  volume={10},
  number={104},
  pages={eadt5254},
  year={2025},
  publisher={American Association for the Advancement of Science}
}

@article{kim2024surgical,
  title={\href{https://arxiv.org/pdf/2407.12998}{Surgical robot transformer (srt): Imitation learning for surgical tasks}},
  author={Kim, Ji Woong and Zhao, Tony Z and Schmidgall, Samuel and Deguet, Anton and Kobilarov, Marin and Finn, Chelsea and Krieger, Axel},
  journal={arXiv preprint arXiv:2407.12998},
  year={2024}
}

@article{reuss2024efficient,
  title={\href{https://arxiv.org/pdf/2412.12953}{Efficient diffusion transformer policies with mixture of expert denoisers for multitask learning}},
  author={Reuss, Moritz and Pari, Jyothish and Agrawal, Pulkit and Lioutikov, Rudolf},
  journal={arXiv preprint arXiv:2412.12953},
  year={2024}
}

@inproceedings{song2024germ,
  title={\href{https://arxiv.org/pdf/2403.13358}{Germ: A generalist robotic model with mixture-of-experts for quadruped robot}},
  author={Song, Wenxuan and Zhao, Han and Ding, Pengxiang and Cui, Can and Lyu, Shangke and Fan, Yaning and Wang, Donglin},
  booktitle={2024 IEEE/RSJ International Conference on Intelligent Robots and Systems (IROS)},
  pages={11879--11886},
  year={2024},
  organization={IEEE}
}

@article{black2024pi0,
  title={\href{https://arxiv.org/pdf/2410.24164}{$\pi_0$: A Vision-Language-Action Flow Model for General Robot Control}},
  author={Black, Kevin and others},
  journal={arXiv preprint arXiv:2410.24164},
  year={2024}
}

@article{octo2024,
  title={\href{https://arxiv.org/pdf/2405.12213}{Octo: An Open-Source Generalist Robot Policy}},
  author={{Octo Model Team} and others},
  journal={arXiv preprint arXiv:2405.12213},
  year={2024}
}

@article{kim2024openvla,
  title={\href{https://arxiv.org/pdf/2406.09246}{OpenVLA: An Open-Source Vision-Language-Action Model}},
  author={Kim, Moo Jin and others},
  journal={arXiv preprint arXiv:2406.09246},
  year={2024}
}

@article{openx2024,
  title={\href{https://ieeexplore.ieee.org/stamp/stamp.jsp?tp=&arnumber=10611477}{Open X-Embodiment: Robotic Learning Datasets and RT-X Models}},
  author={{Open X-Embodiment Collaboration}},
  journal={arXiv preprint arXiv:2310.08864},
  year={2024}
}

@inproceedings{desai2020ablation,
  title={\href{https://ieeexplore.ieee.org/stamp/stamp.jsp?tp=&arnumber=9093360}{Ablation-CAM: Visual explanations for deep convolutional network via gradient-free localization}},
  author={Desai, Saurabh and Ramaswamy, Harish G},
  booktitle={Proceedings of the IEEE/CVF Winter Conference on Applications of Computer Vision},
  pages={983--991},
  year={2020}
}

@article{perera2021global,
  title={\href{https://www.sciencedirect.com/science/article/pii/S1470204520306756}{Global demand for cancer surgery and an estimate of the optimal surgical and anaesthesia workforce between 2018 and 2040: a population-based modelling study}},
  author={Perera, Sathira Kasun and Jacob, Susannah and Wilson, Brooke E and Ferlay, Jacques and Bray, Freddie and Sullivan, Richard and Barton, Michael},
  journal={The Lancet Oncology},
  volume={22},
  number={2},
  pages={182--189},
  year={2021},
  publisher={Elsevier}
}

@article{long2025surgical,
  title={\href{https://pubmed.ncbi.nlm.nih.gov/40668896/}{Surgical embodied intelligence for generalized task autonomy in laparoscopic robot-assisted surgery}},
  author={Long, Yonghao and Lin, Anran and Kwok, Derek Hang Chun and Zhang, Lin and Yang, Zhenya and Shi, Kejian and Song, Lei and Fu, Jiawei and Lin, Hongbin and Wei, Wang and others},
  journal={Science Robotics},
  volume={10},
  number={104},
  pages={eadt3093},
  year={2025},
  publisher={American Association for the Advancement of Science}
}

@inproceedings{collaborative2018perioperative,
  title={\href{https://www.afpp.org.uk/wp-content/uploads/sfa-position-statement-final-april-2018.pdf}{The perioperative care collaborative position statement: surgical first assistant}},
  author={Collaborative, Perioperative Care},
  year={2018},
  organization={PCC}
}

@article{younis2024surgical,
  title={\href{https://link.springer.com/article/10.1007/s00464-024-10958-w}{A surgical activity model of laparoscopic cholecystectomy for co-operation with collaborative robots}},
  author={Younis, R and Yamlahi, A and Bodenstedt, S and Scheikl, PM and Kisilenko, A and Daum, M and Schulze, A and Wise, PA and Nickel, F and Mathis-Ullrich, F and others},
  journal={Surgical Endoscopy},
  volume={38},
  number={8},
  pages={4316--4328},
  year={2024},
  publisher={Springer}
}

@article{chiu2008role,
  title={\href{https://pubmed.ncbi.nlm.nih.gov/18757384/}{The role of the assistant in laparoscopic surgery: important considerations for the apprentice-in-training}},
  author={Chiu, Anita and Bowne, Wilbur B and Sookraj, Kelley A and Zenilman, Michael E and Fingerhut, Abe and Ferzli, George S},
  journal={Surgical innovation},
  volume={15},
  number={3},
  pages={229--236},
  year={2008},
  publisher={SAGE Publications Sage CA: Los Angeles, CA}
}

@article{konstantinidis2020trends,
  title={\href{https://pubmed.ncbi.nlm.nih.gov/31820161/}{Trends and outcomes of robotic surgery for gastrointestinal (GI) cancers in the USA: maintaining perioperative and oncologic safety}},
  author={Konstantinidis, Ioannis T and Ituarte, Philip and Woo, Yanghee and Warner, Susanne G and Melstrom, Kurt and Kim, Jae and Singh, Gagandeep and Lee, Byrne and Fong, Yuman and Melstrom, Laleh G},
  journal={Surgical Endoscopy},
  volume={34},
  number={11},
  pages={4932--4942},
  year={2020},
  publisher={Springer}
}

@article{schussler2025semi,
  title={\href{https://ieeexplore.ieee.org/document/11027660}{Semi-Autonomous Robotic Assistance for Gallbladder Retraction in Surgery}},
  author={Sch{\"u}{\ss}ler, Alexander and Kunz, Christian and Younis, Rayan and Alt, Benjamin and Paik, Jamie and Wagner, Martin and Mathis-Ullrich, Franziska},
  journal={IEEE Robotics and Automation Letters},
  year={2025},
  publisher={IEEE}
}

@article{haworth2025suturebot,
  title={\href{https://suturebot.github.io/static/SutureBot_NeurIPS_2025.pdf}{SutureBot: A Precision Framework \& Benchmark For Autonomous End-to-End Suturing}},
  author={Haworth, Jesse and Chen, Juo-Tung and Nelson, Nigel and Kim, Ji Woong and Moghani, Masoud and Finn, Chelsea and Krieger, Axel},
  journal={arXiv preprint arXiv:2510.20965},
  year={2025}
}

@article{openhelp,
  author = {Kenngott, H. G. and W{\"u}nscher, J. J. and Wagner, M. and Preukschas, A. and Wekerle, A. L. and Neher, P. and Suwelack, S. and Speidel, S. and Nickel, F. and Oladokun, D. and Maier-Hein, L. and Dillmann, R. and Meinzer, H. P. and M{\"u}ller-Stich, B. P.},
  title = {\href{https://pubmed.ncbi.nlm.nih.gov/25673345/}{OpenHELP (Heidelberg laparoscopy phantom): Development of an open-source surgical evaluation and training tool}},
  journal = {Surgical Endoscopy},
  volume = {29},
  number = {11},
  pages = {3338--3347},
  year = {2015},
  doi = {10.1007/s00464-015-4094-0}
}

@inproceedings{brohan2023rt1,
    title={\href{https://www.roboticsproceedings.org/rss19/p025.pdf}{{RT-1}: Robotics Transformer for Real-World Control at Scale}},
    author={Brohan, Anthony and Brown, Noah and Carbajal, Justice and others},
    booktitle={Robotics: Science and Systems (RSS)},
    year={2023}
}

\end{document}